\documentclass[runningheads]{llncs}
\usepackage[T1]{fontenc}
\usepackage{graphicx,verbatim}
\usepackage{multirow}
\usepackage{amssymb}
\usepackage{xcolor}
\usepackage{makecell}
\usepackage{url}
\usepackage[acronym]{glossaries}
\usepackage{amsmath}

\newacronym{brea-depth}{BREA-Depth}{Bronchoscopy REalistic Airway-geometric Depth Estimation}

\begin{document}
\title{BREA-Depth: Bronchoscopy Realistic Airway-geometric Depth Estimation}

\author{Francis Xiatian Zhang\inst{1,2}\orcidID{0000-0003-0228-6359} \and
Emile Mackute\inst{2}\orcidID{0009-0004-4535-4619} \and
Mohammadreza Kasaei\inst{1,2}\orcidID{0000-0003-4932-1457} \and
Kevin Dhaliwal\inst{1}\orcidID{0000-0002-3925-3174} \and
Robert Thomson\inst{3}\orcidID{0000-0003-4978-1488} \and
Mohsen Khadem\inst{1,2}\orcidID{0000-0002-6873-273X}
}
%index{Zhang, Francis Xiatian}
%index{Mackute, Emile}
%index{Kasaei, Mohammadreza}
%index{Dhaliwal, Kevin}
%index{Thomson, Robert}
%index{Khadem, Mohsen}

\authorrunning{Zhang et al.}
\institute{Baillie Gifford Pandemic Science Hub, Institute of Regeneration and Repair, University of Edinburgh, Edinburgh, United Kingdom \and
School of Informatics, University of Edinburgh, Edinburgh, United Kingdom \\
\email{\{francis.zhang, s1657385, m.kasaei, kev.dhaliwal, mohsen.khadem\}@ed.ac.uk}\\ \and
Institute of Photonics and Quantum Science, Heriot-Watt University, Edinburgh, United Kingdom \\
\email{r.r.thomson@hw.ac.uk}}

\maketitle              % typeset the header of the contribution
\begin{abstract}
Monocular depth estimation in bronchoscopy can significantly improve real-time navigation accuracy and enhance the safety of interventions in complex, branching airways. Recent advances in depth foundation models have shown promise for endoscopic scenarios, yet these models often lack anatomical awareness in bronchoscopy, overfitting to local textures rather than capturing the global airway structure—particularly under ambiguous depth cues and poor lighting.
To address this, we propose Brea-Depth, a novel framework that integrates airway-specific geometric priors into foundation model adaptation for bronchoscopic depth estimation. Our method introduces a depth-aware CycleGAN, refining the translation between real bronchoscopic images and airway geometries from anatomical data, effectively bridging the domain gap. In addition, we introduce an airway structure awareness loss to enforce depth consistency within the airway lumen while preserving smooth transitions and structural integrity. By incorporating anatomical priors, Brea-Depth enhances model generalization and yields more robust, accurate 3D airway reconstructions. To assess anatomical realism, we introduce Airway Depth Structure Evaluation, a new metric for structural consistency.
We validate BREA-Depth on a collected ex-vivo human lung dataset and an open bronchoscopic dataset, where it outperforms existing methods in anatomical depth preservation.

% Authors must provide keywords and are not allowed to remove this Keyword section.

\end{abstract}
\section{Introduction}
% BDE is a critical task
% Current DE tech in general CV cannot solve 
% Challenge: Current work still ingore the airway prior
% Solution: 1. CycleGAN between geometric shape and real RGB. 2. Airway loss 

% However, acquiring accurate depth ground truth in bronchoscopy remains challenging. 
%These models leverage self-training strategies to iteratively refine depth predictions, overcoming the constraints of traditional synthetic data. 
% These synthetic depths are typically binary (white airway walls vs.\ black voids), capturing only the lumen surface while omitting finer structural details. 
% As a result, networks that predict more continuous or nuanced depth values are penalized under standard pixelwise metrics, limiting their generalization to diverse bronchoscopic scenarios. 
% This limitation can significantly impact high-precision tasks such as targeted biopsy or localized drug delivery, which require precise anatomical localization within the airway.

Accurate bronchoscopic depth estimation is essential for navigation and 3D airway reconstruction, but obtaining reliable ground truth remains challenging.
Many existing methods~\cite{visentini2017deep,karaoglu2021adversarial,banach2021visually,yang2024adversarial,liu2024robust,guo2024cgan} rely on synthetic depth from CT scans, producing oversimplified, binary depth maps that lack realistic variations and generalizability. As a result, models trained on such data struggle with real-world bronchoscopic conditions.
Recent depth foundation models, such as Depth Anything~\cite{yang2024depth}, offer improved generalization by learning robust depth representations from large-scale unlabeled image sets. 
Inspired by their success, recent works~\cite{cui2024endodac,tian2024endoomni,cui2024surgical} have adapted foundation models for endoscopic applications. However, these methods still prioritize pixel-level accuracy over anatomical awareness, limiting their reliability for bronchoscopic tasks that require structurally precise depth, such as targeted biopsy or localized drug delivery.
%Inspired by their success, recent works~\cite{cui2024endodac,tian2024endoomni,cui2024surgical} have explored transferring foundation-model knowledge to endoscopic applications.

A key challenge in adapting foundation models to bronchoscopy is their \emph{lack of anatomical awareness}. Unlike general endoscopic or open-world scenes~\cite{ryan2017anatomical}, bronchoscopy involves navigating structured yet deformable airway branches. Models relying solely on photometric cues often fail to capture the global airway geometry, particularly under ambiguous depth conditions~\cite{liu2019dense}. Furthermore, large-scale models such as Depth Anything~\cite{yang2024depth} have limited exposure to airway branching topologies, leading to overfitting on local textures rather than learning structural priors. Additionally, standard evaluation metrics (e.g., Abs Rel~\cite{dong2022towards}) focus on pixel-level accuracy but fail to assess global anatomical consistency—an essential feature for navigation and 3D airway reconstruction.

To overcome existing limitations, we propose Bronchoscopy REalistic Airway-geometric Depth Estimation (BREA-Depth), a framework integrating airway-specific structural priors into depth estimation. Our approach employs a \emph{Depth-aware CycleGAN} to refine synthetic-to-real translation, bridging the gap between simulated training data and real bronchoscopic footage. Additionally, we introduce an \emph{Airway Structure Awareness Loss} to enforce depth consistency within the airway lumen while preserving smooth transitions and geometric integrity. To evaluate anatomical consistency, we propose an \emph{Airway Depth Structure Evaluation} metric, assessing depth distribution relative to the airway lumen. For validation, We collected bronchoscopic video data from ex-vivo human lung models and annotated 3,437 bronchoscopic images with semantic segmentation across five navigation sequences.

We extensively evaluate BREA-Depth against state-of-the-art foundation models~\cite{yang2024depth,yang2024depthv2} and recent bronchoscopy-adapted approaches~\cite{cui2024endodac,tian2024endoomni}. Our experiments on both our collected dataset and the open dataset~\cite{visentini2017deep} demonstrate that BREA-Depth significantly outperforms existing techniques, particularly in anatomically complex regions with limited depth cues. The \textbf{source code and ex-vivo human lung dataset} can be found at \url{https://github.com/SIRGLab/BREA-Depth}.

%(\textcolor{blue}{\url{https://anonymous.4open.science/r/BREA-Depth-0A7F}})

Our main contributions are summarized as follows:  
\begin{enumerate}  
    \item BREA-Depth, a novel framework that integrates airway geometry into foundation depth estimation, ensuring anatomically consistent depth predictions while preserving the complex bronchial structure.
    \item We introduce \emph{Airway Depth Structure Evaluation}, a new metric assessing anatomical consistency in depth predictions relative to the airway lumen, complementing existing evaluation methods.  
    \item We open-source a human lung bronchoscopy dataset to support future research on depth estimation in bronchoscopy, facilitating further advancements in airway navigation and intervention.  
\end{enumerate}

\begin{figure}[t]
    \centering
    \includegraphics[width=0.7\columnwidth]{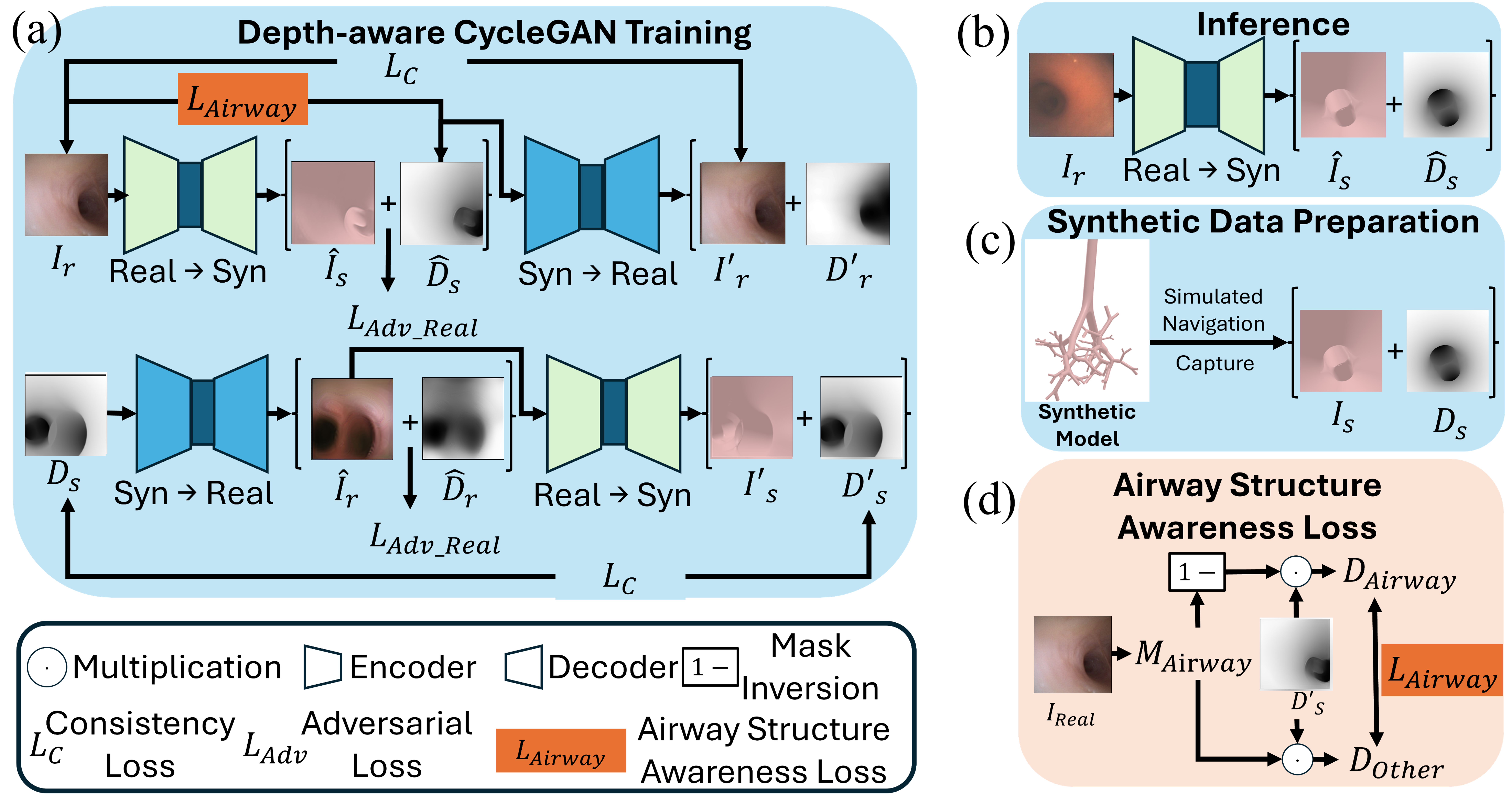}
    \caption{(a) Overview of our bronchoscopic depth estimation framework using a cycle-consistent adversarial approach with unpaired synthetic and real bronchoscopic images through two branches: (i) \emph{Syn-to-Real} translates simulated depth maps and frames from a synthetic airway model into a realistic style, and (ii) \emph{Real-to-Syn} reduces texture artifacts in real images. A foundation model provides pseudo-depth supervision, while a discriminator enforces realism. (b) During inference, a bronchoscopic frame $I_r$ generates a depth map $\hat{D}$ and a synthetic frame $\hat{I}$ that reduces texture noise while preserving geometric details. (c) Geometrically accurate bronchial tree generated in Blender and used to collect synthetic data. (d) Our proposed Airway Structure Awareness loss enforcing depth consistency and smooth transitions within the airway lumen.}
    \label{fig:overview}
\end{figure}

\section{Methodology}

The BREA-Depth model (Fig.~\ref{fig:overview}) processes a bronchoscopic frame \( X \in \mathbb{R}^{H \times W \times 3} \) to generate a depth map \( \hat{D} \in \mathbb{R}^{H \times W \times 1} \) and a synthetic-style frame \( \hat{Y} \in \mathbb{R}^{H \times W \times 3} \), which reduces texture noise while preserving airway geometry. Training follows a CycleGAN approach~\cite{zhu2017unpaired}, leveraging unpaired synthetic but geometrically accurate images and real bronchoscopic images through two branches: \emph{(i) Syn-to-Real}, which refines synthetic depth maps and frames into realistic representations, and \emph{(ii) Real-to-Syn}, which enhances real frames while preserving geometric features. Pseudo ground truth from a foundation model supervises real-depth predictions, while a PatchGAN discriminator~\cite{isola2017image} enforces alignment between generated depth maps and frames. 

The following sections describe our simulation environment for generating geometrically accurate synthetic airway images, the Depth-Aware CycleGAN for depth estimation, and the proposed loss function that enforces depth consistency and smooth transitions within the airway lumen.

\subsection{Geometrically Accurate Model of Airways}
\label{section_airway_model}
We developed a physiologically realistic bronchoscopy simulation by generating 3D lung structures based on key geometric properties: 
(i) airway branching, (ii) bifurcation structure, and (iii) carinal rounding. These parameters are derived from anatomical airway models~\cite{raabe,florens} and airflow-based geometric models~\cite{hegedus,heistracher} to ensure accurate lung morphology representation.

The bronchial tree is modeled as a series of cylindrical airways undergoing binary branching, where each airway has a length \( L \) and diameter \( D \), with the length-to-diameter ratio \( L/D \) specified per generation. Given a parent airway of diameter \( D_p \), daughter branches have diameters \( D_a \) and \( D_b \), defined by \( h_a = D_a / D_p \) and \( h_b = D_b / D_p \), where \( h_a, h_b \in (0,1) \) follow anatomical scaling laws.

The bifurcation region smoothly transitions between the parent and daughter branches, governed by branching angles \( \Phi_a, \Phi_b \), curvature radii \( R^*_a, R^*_b \), and ring radii \( R_a(\phi_{sa}) \), \( R_b(\phi_{sb}) \). The branching angles are sampled within a physiologically valid range \( \Phi_a, \Phi_b \in [\Phi_{\text{min}}, 120^\circ] \), where \( \Phi_{\text{min}} \) prevents airway intersection. The curvature radii are computed as:
\begin{equation}
    R^*_a = \frac{D_a}{2 \sin{\Phi_a}}, \quad R^*_b = \frac{D_b}{2 \sin{\Phi_b}}
\end{equation}
A sigmoid-based transition function ensures gradual bifurcation tapering:  
\( R_a(\phi_{sa}) = D_a \cdot f(\phi_{sa}) \), \( R_b(\phi_{sb}) = D_b \cdot f(\phi_{sb}) \).

Carinal rounding smooths bifurcation transitions using continuous rounding circles centered at \( K \), with radius \( R_c = \min\limits_{(x,y) \in \mathbb{R}^2} \sqrt{(x - K_x)^2 + (y - K_y)^2} \), where \( K_x, K_y \) define the circle centers, and radii are computed for sagittal angles \( \phi_{sa} \in [0, R_a] \) and \( \phi_{sb} \in [0, R_b] \). Unlike previous models assuming parallel circles~\cite{hegedus}, our method dynamically tilts them to match airway asymmetry.

To enhance realism, additional constraints are applied. Twist angles \( \Theta \) are sampled uniformly as \( \Theta \sim \mathcal{U}(0^\circ, 360^\circ) \), while airway lengths follow a Gaussian distribution \( L \sim \mathcal{N}(L_{\text{mean}}, 0.3 L_{\text{mean}}) \), where \( L_{\text{mean}} \) is the expected length per generation. These constraints ensure anatomically plausible airway structures while introducing natural variability. The model is implemented in Blender~\cite{blender} for realistic bronchoscopy simulations, as shown in Fig.~\ref{fig:overview}c.

\subsection{Depth-Aware CycleGAN}
Our simulated depth maps and frames are generated during navigation through the airway model described in the previous section. While synthetic images capture lumen geometry, they lack the texture and structural variations of real bronchoscopic images. To bridge this domain gap, we introduce a \emph{Depth-aware CycleGAN}, refining synthetic-to-real translation while enforcing airway-specific depth constraints. Unlike conventional style transfer methods that rely on feature distribution similarity~\cite{karaoglu2021adversarial,yang2024adversarial}—often failing to preserve structural geometry in bronchoscopy - our approach directly integrates depth into domain translation, enabling more anatomically accurate synthetic-to-real adaptation.

Our framework comprises two U-Net Transformer-like \cite{petit2021u} encoder-decoder branches: \emph{Syn-to-Real}, which refines depth maps to enhance realistic texture and structural variations, and \emph{Real-to-Syn}, which converts real bronchoscopic images into synthetic-style RGB frames and generates depth map, reducing texture noise while preserving geometric features. The translation process follows:

\noindent \emph{Synthetic-to-Real Translation:} Given a synthetic image $X_s = \{I_s, D_s\}$, where $I_s \in \mathbb{R}^{H \times W \times 3}$ represents the synthetic RGB image and $D_s \in \mathbb{R}^{H \times W \times 1}$ is the perfect depth map, the encoder maps it to a latent representation $Z_s$ and the decoder then generates the translated real-style RGB image $\hat{I}_r$ and its corresponding real-style depth $\hat{D}_r$:
\begin{equation}
    Z_s = E_s(D_s), \quad \hat{Y}_s = D_r(Z_s), \quad \hat{Y}_s = \{\hat{I}_r, \hat{D}_r\}
\end{equation}

\noindent \emph{Real-to-Synthetic Translation:} For a real bronchoscopic image $X_r = \{I_r\}$, the encoder extracts its feature representation $Z_r$, and the decoder then predicts the corresponding synthetic perfect depth $\hat{D}_s$ while reconstructing the synthetic-style RGB image $\hat{I}_s$:
\begin{equation}
    Z_r = E_r(X_r), \quad \hat{Y}_r = D_s(Z_r), \quad \hat{Y}_r = \{\hat{I}_s, \hat{D}_s\}
\end{equation}
Similarly, in another cycle, based on the generated $\{\hat{I}_r, \hat{D}_r\}$ and $\{\hat{I}_s, \hat{D}_s\}$, respectively, our framework further generates the synthetic $\{I'_s, D'_s\}$ and real domain $\{I'_r, D'_r\}$, which are then used to compute the consistency loss.

\subsection{Airway Structure Awareness Loss}
Unlike general endoscopic or natural scenes, where depth cues rely on texture variations~\cite{ming2021deep,yang20243d}, bronchoscopic images have a monotonous lumen texture with low saliency~\cite{karaoglu2021adversarial}, making pixel-wise depth learning prone to errors. A key anatomical prior in bronchoscopy is that airway depth should increase as the lumen extends further into the respiratory system~\cite{amador2023anatomy}, yet standard depth estimation models fail to enforce this, leading to inconsistent predictions. To address this, we introduce the \emph{Airway Structure Loss}, which encourages lower disparity values in the airway lumen by leveraging intensity-based segmentation.

The airway region is defined using a grayscale threshold \( T \) on the grayscale-translated image \( I_{\text{gray}} \), forming a binary mask \( M_{\text{airway}} \) that identifies the airway lumen
    $M_{\text{airway}} = (I_{\text{gray}} < T).$
The mean disparity in airway and non-airway regions is computed as:
\begin{equation}
    D_{\text{airway}} = \frac{\sum (\hat{D} \cdot M_{\text{airway}})}{\sum M_{\text{airway}} + \epsilon}, \quad 
    D_{\text{non-airway}} = \frac{\sum (\hat{D} \cdot (1 - M_{\text{airway}}))}{\sum (1 - M_{\text{airway}}) + \epsilon}.
\end{equation}
To enforce anatomical consistency, we use a ReLU-based formulation to penalize cases where the airway disparity exceeds the non-airway disparity:
\begin{equation}
    \mathcal{L}_{\text{airway}} = \mathbb{E} \left[ \max(0, D_{\text{airway}} - D_{\text{non-airway}}) \right].
\end{equation}

The total optimization objective is:
\begin{equation}
    \mathcal{L}_{\text{total}} = \lambda_{\text{adv}} \mathcal{L}_{\text{adv}} + \lambda_{\text{cycle}} \mathcal{L}_{\text{cycle}} + \lambda_{\text{identity}} \mathcal{L}_{\text{identity}} + \lambda_{\text{airway}} \mathcal{L}_{\text{airway}},
\end{equation}
where \( \mathcal{L}_{\text{adv}} \) is the adversarial loss for realistic image translation~\cite{isola2017image}, \( \mathcal{L}_{\text{cycle}} \) ensures bidirectional cycle consistency, \( \mathcal{L}_{\text{identity}} \) preserves identity mappings, and \( \mathcal{L}_{\text{airway}} \) enforces depth consistency within the airway lumen. \( \mathcal{L}_{\text{cycle}} \) and \( \mathcal{L}_{\text{identity}} \) are computed using L1 loss.

\section{Experiment}
\begin{figure}[t]
    \centering
    \includegraphics[width=0.8\columnwidth]{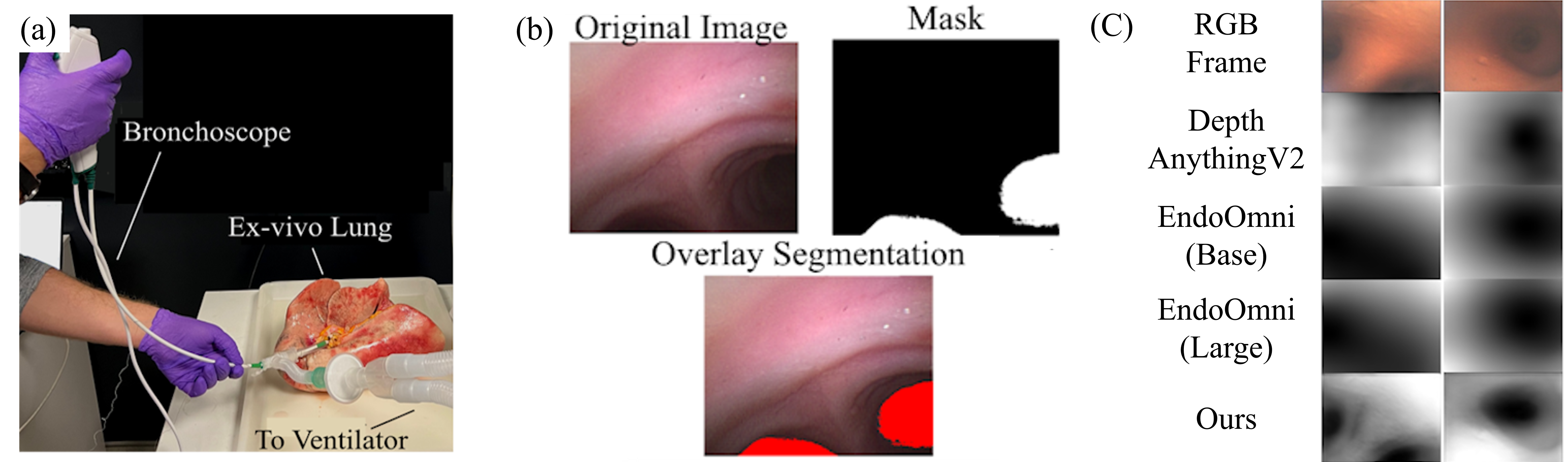}
    \caption{(a) The experimental setup for data collection with ex-vivo human lung. A commercial bronchoscope (Ambu aScope 4 ,Ambu Ltd.) is used by an expert bronchoscopist to navigate the lung. The collection fits the appropriate institutional ethical process. (b) Airway lumen segmentation from our dataset, providing essential ground truth for \emph{Airway Depth Structure Evaluation}. (c) Qualitative comparison of our model's depth estimation results (disparity). The predicted depth map clearly highlights the airway lumen and captures fine anatomical details around airway branches.}
    \label{fig:seg}
\end{figure}

Our model is trained on an NVIDIA RTX 3080 using PyTorch, and utilizes 9,500 synthetic image-depth pairs (Section~\ref{section_airway_model}) and 55,000 real bronchoscopic images (Fig.~\ref{fig:seg}a) with pseudo-depth from DepthAnything~\cite{yang2024depth}. Training runs for 30 epochs with a batch size of 2 and a learning rate of 0.0001.
The weighting factors are set as follows: $\lambda_{\text{adv}} = 5$, $\lambda_{\text{cycle}} = 1$, $\lambda_{\text{identity}} = 1$, and $\lambda_{\text{airway}} = 0.5$.  During inference, our model runs at 60 FPS, achieving online real-time performance.

We compare our framework with existing depth estimation methods, including bronchoscopy-specific approaches and general foundation models:  
1) 3cGAN~\cite{banach2021visually}: A CycleGAN-based method for manually rendering synthetic images.
2) DepthAnything~\cite{yang2024depth}: A foundation model trained on large-scale unlabeled data for monocular depth estimation.  
3) DepthAnythingV2~\cite{yang2024depthv2}: An improved version of DepthAnything.  
4) EndoDAC~\cite{cui2024endodac}: An adaptation of DepthAnything for endoscopy.  
5) EndoOmni~\cite{tian2024endoomni}: An adaptation of DepthAnything leveraging annotated data, with an extended dataset for bronchoscopy.
For foundation models, we use their largest versions, but for EndoOmni, we evaluate both the base (${B}$) and large (${L}$) versions for a comprehensive comparison.

We evaluate depth prediction using two evaluation strategies:

\noindent \textbf{(i) Airway Depth Structure Evaluation:}
Standard depth metrics prioritize pixel accuracy~\cite{dong2022towards} but often neglect anatomical consistency in bronchoscopic navigation~\cite{banach2021visually}. We propose a depth structure evaluation assuming the airway lumen is the deepest region in bronchoscopic imagery~\cite{amador2023anatomy}, assessing: 1) the localization of the lowest depth regions and 2) the depth contrast between lumen and non-lumen regions.

\noindent \emph{Lowest Depth Localization Accuracy (LocalAccu):}  
A well-trained model should predict the lowest depth within the lumen, not along airway walls. Given a predicted depth map \( D \) and airway mask \( M_{\text{lumen}} \), the minimum depth value is $D_{\min} = \min(D)$.
To assess alignment with the airway lumen, we compute the proportion of pixels where \( D = D_{\min} \) within \( M_{\text{lumen}} \):

\begin{equation}
    R_{\text{in-lumen}} = \frac{\sum \mathbb{I}(D = D_{\min}) \odot M_{\text{lumen}}}{\sum \mathbb{I}(D = D_{\min}) + \epsilon}.
\end{equation}
Here, \( M_{\text{lumen}} \) is a binary mask indicating the airway lumen, \( \mathbb{I}(\cdot) \) is an indicator function, and \( \epsilon \) prevents division by zero. If \( R_{\text{in-lumen}} > 0.99 \), the model is considered to have correctly localized the lowest depth inside the lumen.

\noindent \emph{Depth Contrast Consistency (DepthCon):}  
We also assess whether the predicted depth distribution reflects airway structure by computing the mean depth inside and outside the lumen as 
$\bar{D}_{\text{lumen}} = \sum D \odot M_{\text{lumen}} / (\sum M_{\text{lumen}} + \epsilon)$ and 
$\bar{D}_{\text{outside}} = \sum D \odot (1 - M_{\text{lumen}}) / (\sum (1 - M_{\text{lumen}}) + \epsilon)$, respectively. To quantify the contrast, we compute the z-score:
\begin{equation}
    Z_{\text{lumen-outside}} = \frac{\bar{D}_{\text{lumen}} - \bar{D}_{\text{outside}}}{\sigma_{\text{outside}} + \epsilon},
\end{equation}
where \( \sigma_{\text{outside}} \) is the depth standard deviation outside the lumen. A sufficiently negative \( Z_{\text{lumen-outside}}\) (i.e., \( < -1.00 \)) confirms the lumen is significantly deeper.

To address the lack of a dedicated dataset for this evaluation, we create and will open source a segmentation dataset based on five navigation sequences in an ex-vivo human lung (Fig.~\ref{fig:seg}a), comprising over 20 minutes of data recorded at 2.5 fps, totaling 3,437 frames. Each frame is manually prompted and automatically segmented using SegmentAnythingV2~\cite{ravi2024sam} to identify the airway lumen (Fig.~\ref{fig:seg}b).

\noindent \textbf{(ii) Classic Depth Performance Metric:} We benchmark our framework on the Visentini-Scarzanella et al.~\cite{visentini2017deep} dataset, comprising 16 videos (39,599 frames) of a bronchial phantom with ground-truth depth and CT renderings. Following prior work~\cite{cui2024endodac,tian2024endoomni}, we evaluate performance using median alignment post-processing between predicted depth and ground truth, followed by standard depth metrics: Abs Rel, Sq Rel, RMSE, RMSE log, and threshold accuracy ($\delta$).

\begin{table}[h]
\centering
{\fontsize{8pt}{9pt}\selectfont
\caption{Comparison of Airway Depth Structure Evaluation on our collected dataset and Classical Performance on the phantom dataset~\cite{visentini2017deep}. DA: Depth Anything. w/o CycleGAN: Supervised directly using real images with pseudo-depth and synthetic images with depth. w/o $\mathcal{L}{\text{airway}}$: Excludes $\mathcal{L}{\text{airway}}$ from the optimization process.}
\begin{tabular}{l|cc|ccccc}
\hline
\multirow{2}{*}{Methods} & \multicolumn{2}{c|}{\textbf{Structure Evaluation}} & \multicolumn{5}{c}{\textbf{Classical Performance}} \\
 & DepthCon$\uparrow$  & LocalAccu$\uparrow$  & $Abs$\newline$Rel\downarrow$ & $Sq$\newline$Rel\downarrow$ &  $RMSE\downarrow$ & $RMSE$\newline$_{log}\downarrow$ & $\delta\uparrow$\\
\hline
3cGAN~\cite{banach2021visually} &99.27  &57.00  & 0.33 & 7.79 &15.67 &0.35 &57.84 \\
DA~\cite{yang2024depth} & 70.55 & 45.64 & 0.24 & 4.21 &12.98 &0.27  & 64.25\\
DA-V2~\cite{yang2024depthv2} & 60.88 & 51.70 & 0.21 &3.72 &12.02& 0.25& 68.52\\
EndoDAC~\cite{cui2024endodac} & 34.58  & 14.18 & 0.27  & 5.56 & 14.13& 0.30& 63.31 \\
EndoOmni$_{B}$~\cite{tian2024endoomni} & 96.53 &58.19& 0.18 &2.99  &10.66 &0.21 &75.56 \\
EndoOmni$_{L}$~\cite{tian2024endoomni} & 96.09 & 45.79 & 0.18 &2.77  &10.27 &0.21 &76.54\\
\hline
Ours &97.27  &62.36 & 0.23  &4.56 &12.26 &0.25 &70.64 \\
w/o CycleGAN &  68.36& 25.36 & 0.33  &7.76 &16.16 &0.35 &56.89 \\
w/o $\mathcal{L}_{\text{airway}}$ &96.67  &52.06  & 0.29  &6.12 &14.60 &0.31 &59.23 \\
\hline
\end{tabular}
\label{tab:SOTA}
}
\end{table}
\noindent \textbf{Airway Depth Structure Evaluation}
Table~\ref{tab:SOTA} shows our method achieving the highest performance in \emph{airway depth structure evaluation}, with 97.27\% \emph{Depth Contrast Consistency} and 62.36\% \emph{Lowest Depth Localization Accuracy}, outperforming DepthAnything~\cite{yang2024depth} and EndoOmni~\cite{tian2024endoomni}.  
Our framework enhances anatomical consistency and depth structure preservation, benefiting airway reconstruction. Notably, EndoDAC~\cite{cui2024endodac} performs poorly, underscoring the need for airway-specific priors in foundation models. Fig.~\ref{fig:seg}c illustrates our model’s performance, capturing anatomical details around the lumen and airway branches.

\begin{figure}[t]
    \centering
    \includegraphics[scale = 0.20]{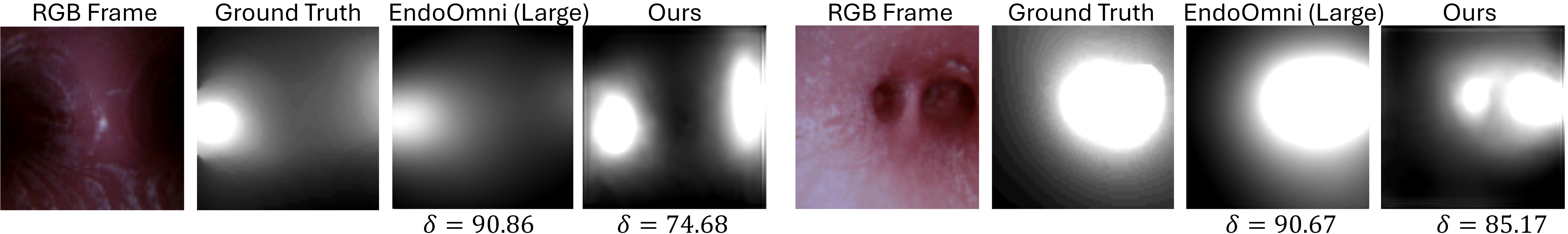}
    \caption{Qualitative comparison of our model's depth estimation (1/disparity) on phantom data~\cite{visentini2017deep}. Our model preserves structural consistency, accurately capturing airway depth, especially at bifurcations. However, pixel-level metrics overlook ground truth inaccuracies (e.g., misassigned depth in extended lumen regions), limiting its apparent performance (e.g., $\delta$) despite improvements.}
    \label{fig:vis_gt}
\end{figure}

\noindent \textbf{Classical Depth Estimation Performance}  
Table~\ref{tab:SOTA} compares our method with existing approaches, showing comparable performance. The limited improvement is largely due to low-quality ground truth from phantom data, which fails to capture real-world bronchoscopic complexity. As shown in Fig.~\ref{fig:vis_gt}, the ground truth often misassigns depth in extended lumen regions or overlooks bifurcations, leading to inaccurate evaluations. Despite this, our method generalizes well, performing comparably to other foundation models.

\noindent \textbf{Ablation Analysis}  
Table~\ref{tab:SOTA} assesses component contributions. Removing CycleGAN (\emph{w/o CycleGAN}) degrades performance, showing its role in domain adaptation. Excluding airway structure awareness loss ($\mathcal{L}_{\text{airway}}$) weakens structural preservation, emphasizing its importance for anatomical consistency. Both components are crucial for achieving anatomically consistent depth predictions.

\section{Conclusion}  
We propose BREA-Depth, a bronchoscopic depth estimation framework that integrates airway-specific geometric priors into foundation model adaptation. By incorporating a \emph{Depth-aware CycleGAN} and an \emph{Airway Structure Awareness Loss}, our approach enhances depth consistency and anatomical realism, outperforming existing methods in structural preservation, as validated by our \emph{Airway Depth Structure Evaluation}.  
Our results highlight the limitations of current evaluation metrics, which emphasize pixel accuracy over anatomical consistency, and the challenges posed by low-quality ground truth in existing datasets. To address this, we introduce new evaluation metrics tailored to bronchoscopic depth estimation.  
Future work will focus on developing higher-quality bronchoscopic depth datasets and integrating depth estimation with bronchoscopy camera pose localization~\cite{deng2023feature} and landmark recognition~\cite{vu2024bm} to improve real-world applicability.

\begin{credits}
\subsubsection{\ackname} This work was supported by the UKRI Medical Research Council (MR/T023252/1), Baillie Gifford Pandemic Science Hub, and the Engineering and Physical Sciences Research Council (EPSRC) ‘U-care’ Programme Grant (EP/T020903/1).

\subsubsection{\discintname}
The authors declare no conflicts of interest in this paper.
\end{credits}

\newpage
\bibliographystyle{splncs04}
\bibliography{ref}
\end{document}